\documentclass[11pt]{article}
\usepackage{preamble}

\title{\fontsize{20pt}{22pt}\selectfont
{\bf Bike2Vec: Vector Embedding Representations of Road Cycling Riders and Races}}

\author{\vspace{8pt}
Ethan Baron\textsuperscript{a}, Bram Janssens\textsuperscript{b,c,d} and Matthias Bogaert\textsuperscript{b,c} \\    
\fontsize{10pt}{6pt}\selectfont    
\textsuperscript{a} University of Toronto / Zelus Analytics, \href{mailto:eth.baron@mail.utoronto.ca}{eth.baron@mail.utoronto.ca}\\   
\fontsize{10pt}{6pt}\selectfont    
\textsuperscript{b} Department of Marketing, Innovation and Organisation, Ghent University\\
\fontsize{10pt}{6pt}\selectfont    
\textsuperscript{c} FlandersMake@UGent–corelab CVAMO \\
\fontsize{10pt}{6pt}\selectfont    
\textsuperscript{d  } Research Foundation Flanders 
}

\date{} 

\begin{document}

\pagenumbering{gobble} 

\maketitle

\vspace{-30pt}

\begin{abstract}
\vspace{-4mm}
\indent Vector embeddings have been successfully applied in several domains to obtain effective representations of non-numeric data which can then be used in various downstream tasks. We present a novel application of vector embeddings in professional road cycling by demonstrating a method to learn representations for riders and races based on historical results. We use unsupervised learning techniques to validate that the resultant embeddings capture interesting features of riders and races. These embeddings could be used for downstream prediction tasks such as early talent identification and race outcome prediction.
\end{abstract}

\section{Introduction}

Professional road cycling offers several interesting challenges in an analytics setting due to its unique properties. Notably, races have a variety of different formats (e.g. one-day races, stage races) and profiles (e.g. flat, hilly, or mountainous), each suiting riders of different characteristics. Although several past works have demonstrated the potential of using machine learning models to predict road cycling race results, these models rely on significant feature engineering efforts and are tailored to predicting specific outcomes, such as rider performance in a specific race.

We present a framework forming the foundation for a generalized prediction algorithm that does not depend on labour-intensive feature engineering efforts. Specifically, we introduce a method to train vector embeddings for riders and races based on historical race results.

In representation learning, vector embeddings are used to capture the key qualities of entities such as words, images, or songs. If trained effectively, these vector embeddings can then be used for a variety of downstream tasks. For example, word embeddings trained using a large corpus of text can be used for emotion recognition or sentence completion.

Likewise, we show that our cycling embeddings capture the key characteristics of riders and races. The embeddings can be used in downstream prediction tasks and eliminate the need for a manual feature engineering process.

\section{Literature Review}

\subsection{Machine Learning in Road Cycling}

There are several prior works which apply machine learning to road cycling.

Multiple works focus on predicting the ProCyclingStats (``PCS'') points, a system developed by the website \url{procyclingstats.com} to assign scores to riders based on the results achieved in certain races. For example, \textcite{Janssens_Bogaert} construct a random forest regression to predict the total PCS points scored by under-23 prospects in their first two years as professional athletes. They engineer a large set of features, including the riders' performances in particular under-23 races, and compare various methods to impute non-participated race results. This imputation method is used to detect the most promising young athletes (\cite{Pogacar}).  Similarly, \textcite{Merckx} compare linear regression and random forest regression to predict the points scored by under-23 riders in their first three years as professionals. They also hand-craft a number of features summarizing the riders' performance at the youth level.

Other works focus on predicting the outcomes of particular races. \textcite{DeSpiegeleerThesis} develops machine learning models to predict various outcomes of stages from the Tour de France, Giro d'Italia, and Vuelta a Espana, including their average speed, the difference between the average speed of the winner and that of a particular rider, and the head-to-head performance between two riders. The predictions are based on an extensive set of engineered features related to the terrain, weather, rider characteristics, and historical results. \textcite{Mortirolo} uses Bayesian Additive Regression Trees to simulate races and obtain predictions for the probabilities of specific race outcomes. The simulation uses over one hundred features, including ratings for riders' various abilities, indicators of riders' recent form, historical results from the past three years, and team-level indicators. \textcite{Kholkine} apply an XGBoost model to predict the outcomes of the Tour of Flanders using riders' performances in relevant build-up races. They also engineer several advanced features related to past results in similar races, historical weather data and overall team performance. \textcite{L2R} also employ an XGBoost model within a learn-to-rank framework to predict the top ten riders in several one-day races using a suite of engineered features based on historical results and the riders' ages. Finally \textcite{DemunterThesis} compares linear regression, random forest, XGBoost, and neural networks to predict the rankings of riders in a given race. Again, various features related to the rider's recent and historical results are developed and used as inputs for these models.

To the best of our knowledge, we present the first framework for a generalized prediction algorithm for road cycling which does not rely on a hand-crafted set of features for the particular outcome of interest.

\subsection{Representation Learning}

Representation learning is the field of machine learning concerned with automatically learning meaningful and compact representations of data without requiring features for the data. These representations aim to capture the underlying structure and patterns in the data, enabling more effective performance on a variety of downstream applications. Some primary applications of representation learning include natural language processing, computer vision, and recommendation systems.

In natural language processing, representation learning has been utilized to learn word embeddings that capture semantic relationships between words. For example, the word2vec algorithm uses a skip-gram approach to fit vector embeddings for words. These embeddings yield successful results on downstream tasks such as semantic and syntactic word relationship testing and sentence completion (\cite{word2vec}). Another common approach known as GloVe uses co-occurence statistics of words to obtain the vector embeddings. The resulting representations performed strongly on a variety of tasks including word analogies, word similarities, and named entity recognition (\cite{glove}). Finally, pre-trained vector embeddings using bidirectional encoder representations from transformers (BERT) can then be used as inputs into various downstream tasks and have yielded state-of-the-art performance on question answering and language inference (\cite{bert}).

Similarly, representation learning has been successfully applied in the field of computer vision. One notable technique is the use of convolutional neural networks. These convolutional neural networks achieved pioneering performance on image classification tasks, such as handwritten digit recognition (\cite{MNIST}) and high-resolution image classification (\cite{ImageNet}). By pre-training large convolutional neural networks on large amounts of data, researchers have then achieved state-of-the-art performance on novel computer vision tasks, including image recognition (\cite{simonyan2015deep}), object detection (\cite{girshick}), scene recognition and domain adaptation (\cite{donahue2013decaf}).

Representation learning advancements in natural language processing and computer vision have exploited observed relationships between words and local patterns in images. In the case of road cycling, we seek to extract representations for both races and riders and exploit historical interactions between these two types of entities. Most relevant to this context are past works on recommender systems surrounding collaborative filtering, which use historical interactions between users and items to recommend new items to a user. One common approach for such problems is to transform both items and
users to the same vector space by assigning vector embeddings of the same dimension to both categories. Then, dot products between the vector embeddings of users and items can be used to infer their interaction (\cite{colab_filter}).

Although we are not aware of such a representation learning approach being applied in road cycling, similar approaches have recently been tested in other sports, including soccer (\cite{coach2vec, football2vec, yilmaz, li}), basketball (\cite{thoops}), and cricket (\cite{alaka}).

\section{Methods}

We collect historical race results from the 2016-2022 UCI World Tour seasons from \url{procyclingstats.com}. Specifically, we consider the results of one-day races, and individual stages of stage races (i.e. multi-day races), except for team time trials. We do not consider overall classifications of stage races. Overall, our dataset includes results from 1069 race editions, 118 of which are one-day races.

For each result in our data, we define the \emph{normalized PCS score} as the number of PCS points scored by the rider in the race, divided by the points earned by the winner of that race. For example, if the winner and runner-up of a race earn 500 and 300 PCS points respectively, they are assigned a normalized PCS score of 1 and 0.6 respectively.

We learn vector embeddings of dimension $D$ for individual riders and races by directly optimizing these embeddings' ability to predict historical results. Specifically, we represent a rider's aptitude at a given race by the dot-product between that rider's embedding and that race's embedding. We then pass this dot-product through a sigmoid activation function to predict the normalized PCS score for that rider at that race. The vector embeddings are trained by  minimizing the binary cross-entropy loss between these predictions and the actual normalized PCS scores, according to equation \ref{loss_func}.

\begin{equation} \label{loss_func}
    L(R, S, y) = \frac{1}{N} \sum_{i = 1}^N y_i \log(\sigma(R_{r(i)} \cdot S_{s(i)})) + (1 - y_i) \log (1 - \sigma(R_{r(i)} \cdot S_{s(i)}))
\end{equation}

Here, $y_i$ records the normalized PCS points scored by rider $r_{(i)}$ in race $s_{(i)}$, $R$ is the matrix of rider vector embeddings, $S$ is the matrix of race vector embeddings, and $N$ is the number of results in our data. $\sigma$ refers to the sigmoid function such that $\sigma(x) = (1 + e^{-x})^{-1}$.

We train a vector embedding for each rider who has scored at least 25 (unnormalized) PCS points in our dataset. We also train a vector embedding for each race edition, except that for one-day races, we use the same embedding across all seasons. We do this since one-day races tend to suit similar riders across years, whereas stages can have very different characteristics across seasons. In total, we train unique embeddings for 973 races and 958 riders.

The results shown below are based on embeddings of dimension $D = 5$ trained using an Adam optimizer with a learning rate of 0.001 for 100 epochs. Reproducible code to implement our methods is available at \url{https://github.com/baronet2/Bike2Vec}.

\section{Results}

In this section, we analyze our learned embeddings to demonstrate that they capture valuable information about the characteristics of riders and races.

\begin{figure}[h!]
  \centering
      \includegraphics[width=0.8\textwidth]{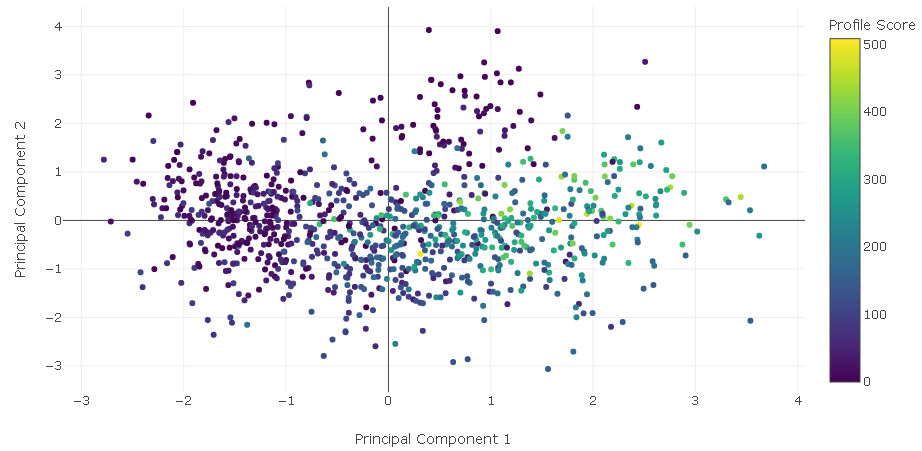}
  \caption{\label{race_pca} Visualization of race embeddings coloured by race profile score.}
\end{figure}

In Figure \ref{race_pca}, we plot the embeddings for each race in our dataset, coloured according to the race profile score, a measurement of the amount of climbing in the race developed by PCS. We performed a principal component analysis to reduce the dimensionality of the embeddings to two dimensions for visualization purposes. Clearly, the primary principal component is capturing significant information about the terrain of a race, with more mountainous races appearing on the right and flat races appearing on the left.

\begin{figure}[h!]
  \centering
      \includegraphics[width=0.8\textwidth]{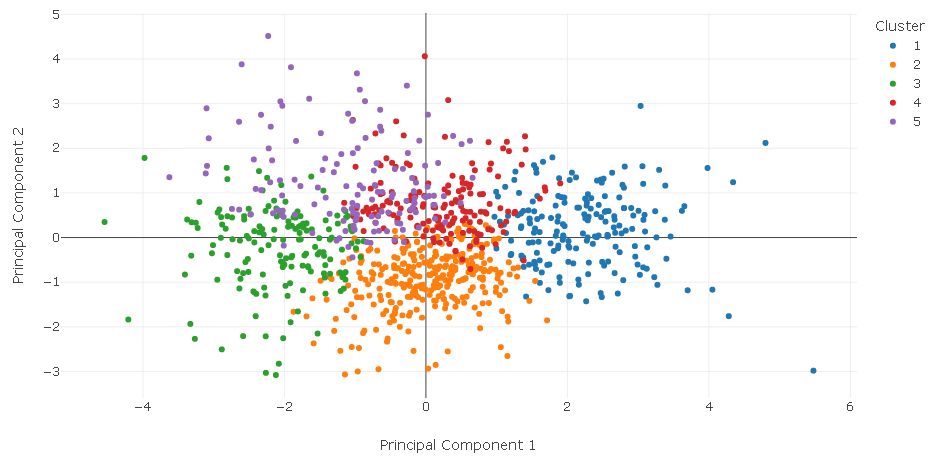}
  \caption{\label{rider_pca} Visualization of rider embeddings coloured by cluster.}
\end{figure}

Similarly, in Figure \ref{rider_pca}, we plot the principal components of the rider embeddings. Unlike races, riders are not labelled by PCS as belonging to a certain category. Therefore, to add interpretability, we perform k-means clustering on the rider embeddings and colour the riders by their assigned cluster.

We show a few examples of riders from each cluster in Table \ref{rider_clusters}. There are clear similarities among the riders in each cluster, indicating that our embeddings are capturing the unique characteristics of each rider. For example, cycling fans would identify that cluster 1 is composed of sprinters and cluster 3 of climbers.

\begin{table}[!h]
\begin{center}
\begin{tabular}{c | l} 
 Cluster & Examples of Riders \\
 \hline\hline
 1 & SAGAN Peter, KRISTOFF Alexander, VIVIANI Elia, EWAN Caleb, BENNETT Sam \\ \hline
 2 & VAN AVERMAET Greg, COLBRELLI Sonny, NAESEN Oliver, MOHORIČ Matej \\ \hline
 3 & ALAPHILIPPE Julian, VALVERDE Alejandro, ROGLIČ Primož, POGAČAR Tadej \\ \hline
 4 & VAN AERT Wout, MATTHEWS Michael, STUYVEN Jasper, KWIATKOWSKI Michał \\ \hline
 5 & VAN DER POEL Mathieu, GILBERT Philippe, LAMPAERT Yves, ŠTYBAR Zdeněk \\ \hline
\end{tabular}
\caption{\label{rider_clusters} Examples of riders from each cluster.}
\end{center}
\end{table}

\begin{table}[!h]
\begin{center}
\begin{tabular}{c | c} 
 Rider 1 & Rider 2 \\
 \hline\hline
 POGAČAR Tadej & ROGLIČ Primož \\ \hline
 SAGAN Peter & COLBRELLI Sonny \\ \hline
 ALAPHILIPPE Julian & HIRSCHI Marc \\ \hline
 YATES Simon & BARDET Romain \\ \hline
 EVENEPOEL Remco & ALMEIDA João \\ \hline
 QUINTANA Nairo & ZAKARIN Ilnur \\ \hline
 VIVIANI Elia & GREIPEL André \\ \hline
 DENNIS Rohan & CAVAGNA Rémi \\ \hline
\end{tabular}
\caption{\label{rider_similarity} Examples of most similar rider embeddings.}
\end{center}
\end{table}

Furthermore, in Table \ref{rider_similarity}, we show some examples of rider similarities. That is, for each rider on the left-hand-side, we show the name of the other rider with the most similar embedding, according to Euclidean distance. Cycling fans would confirm that these rider pairings strongly reflect these riders' characteristics. For example, Tadej Pogacar and Primoz Roglic are both world-class climbers and time-trialists, Peter Sagan and Sonny Colbrelli are versatile sprinters who also perform well in cobbled or hilly classics, and Julian Alaphilippe and Marc Hirschi are both specialists at climbing short but steep hills.

Overall, the vector embeddings seem to accurately capture the distinguishing characteristics of both riders and races.

\section{Conclusion}
We present a novel vector embedding approach to represent road cycling riders and races, and implement the approach on seven seasons of data from professional men's road cycling races. We validate the resulting embeddings by showing that they contain useful information about the characteristics of races and riders. These embeddings can form the basis for a variety of downstream prediction tasks, removing the need for extensive manual feature engineering.

Although we have demonstrated that our proposed vector embeddings contain valid information about the riders and races, we have yet to test the inclusion of these embeddings within a downstream prediction task. We leave this as a promising area for future work. Further, augmenting our race embeddings using features about the route, such as the elevation profile, could offer improved race embeddings and enable predictions on new races. Additionally, our current framework assigns a constant embeddings over the span of riders' careers. Future research could seek to incorporate a time-varying element to capture changes in rider skills due to age, physiology, injury, or other effects. Lastly, an interesting avenue for further exploration is extending our framework to women's cycling and comparing the results.

\newpage

\printbibliography

\end{document}